\documentclass[10pt,twocolumn,letterpaper]{article}

\usepackage{iccv}
\usepackage{times}
\usepackage{epsfig}
\usepackage{graphicx}
\usepackage{amsmath}
\usepackage{amssymb}
\usepackage{bbding}
\usepackage{booktabs}
\usepackage[numbers,sort]{natbib}

\usepackage{soul}
\usepackage[dvipsnames]{xcolor}
\usepackage{xspace}
\usepackage[normalem]{ulem}

\usepackage{tabularx,ragged2e,booktabs}
\usepackage{multirow}

\usepackage{color, colortbl}
\definecolor{Gray}{gray}{0.9}

\usepackage{enumitem}

\usepackage[heightadjust=all,valign=c]{floatrow}
\newfloatcommand{capbtabbox}{table}[][\FBwidth]

\usepackage[outline]{contour}
\usepackage{mathtools}
\usepackage{bbm}
\usepackage[breaklinks=true,bookmarks=false]{hyperref}
\usepackage[capitalize]{cleveref}
\crefname{section}{Sec.}{Secs.}
\Crefname{section}{Section}{Sections}
\Crefname{table}{Table}{Tables}
\crefname{table}{Tab.}{Tabs.}

\usepackage{sg-macros} 

\makeatletter
\renewcommand{\paragraph}{%
  \@startsection{paragraph}{4}%
  {\z@}{1.05ex \@plus 1ex \@minus .2ex}{-1em}%
  {\normalfont\normalsize\bfseries}%
}
\makeatother

\iccvfinalcopy 

\def\iccvPaperID{****} 

\ificcvfinal\fi

\begin{document}

\title{Temporal Consistent Automatic Video Colorization via Semantic Correspondence}

\author{Yu Zhang\textsuperscript{1}, Siqi Chen\textsuperscript{1}\footnotemark[1], Mingdao Wang\textsuperscript{1}, Xianlin Zhang\textsuperscript{2}, Chuang Zhu\textsuperscript{1}, Yue Zhang\textsuperscript{2}, Xueming Li\textsuperscript{2}\\
\textsuperscript{1}School of Artificial Intelligence, Beijing University of Posts and Telecommunications\\
\textsuperscript{2}School of Digital Media and Design Arts, Beijing University of Posts and Telecommunications\\
Beijing, China\\
{\tt\small \{zhangyu\_03, sqchen, wmingdao, zxlin, czhu\}@bupt.edu.cn,}\\
{\tt\small zhangyuereal@163.com, lixm@bupt.edu.cn}}

\maketitle
\renewcommand{\thefootnote}{\fnsymbol{footnote}}
\footnotetext[1]{Corresponding author}

\begin{abstract}
Video colorization task has recently attracted wide attention.
Recent methods mainly work on the temporal consistency in adjacent frames or frames with small interval. However, it still faces severe challenge of the inconsistency between frames with large interval.
To address this issue, we propose a novel video colorization framework, which combines semantic correspondence into automatic video colorization to keep long-range consistency.
Firstly, a reference colorization network is designed to automatically colorize the first frame of each video, obtaining a reference image to supervise the following whole colorization process. 
Such automatically colorized reference image can not only avoid labor-intensive and time-consuming manual selection, but also enhance the similarity between reference and grayscale images. Afterwards, a semantic correspondence network and an image colorization network are introduced to colorize a series of the remaining frames with the help of the reference.
Each frame is supervised by both the reference image and the immediately colorized preceding frame to improve both short-range and long-range temporal consistency. Extensive experiments demonstrate that our method outperforms other methods in maintaining temporal consistency both qualitatively and quantitatively. In the NTIRE 2023 Video Colorization Challenge, our method ranks at the 3rd place in Color Distribution Consistency (CDC) Optimization track.  Code will be available online at \url{https://github.com/bupt-ai-cz/TCVC}.
\end{abstract}


\section{Introduction}
\begin{figure*}[!t]
   \centering
   \includegraphics[width=\linewidth]{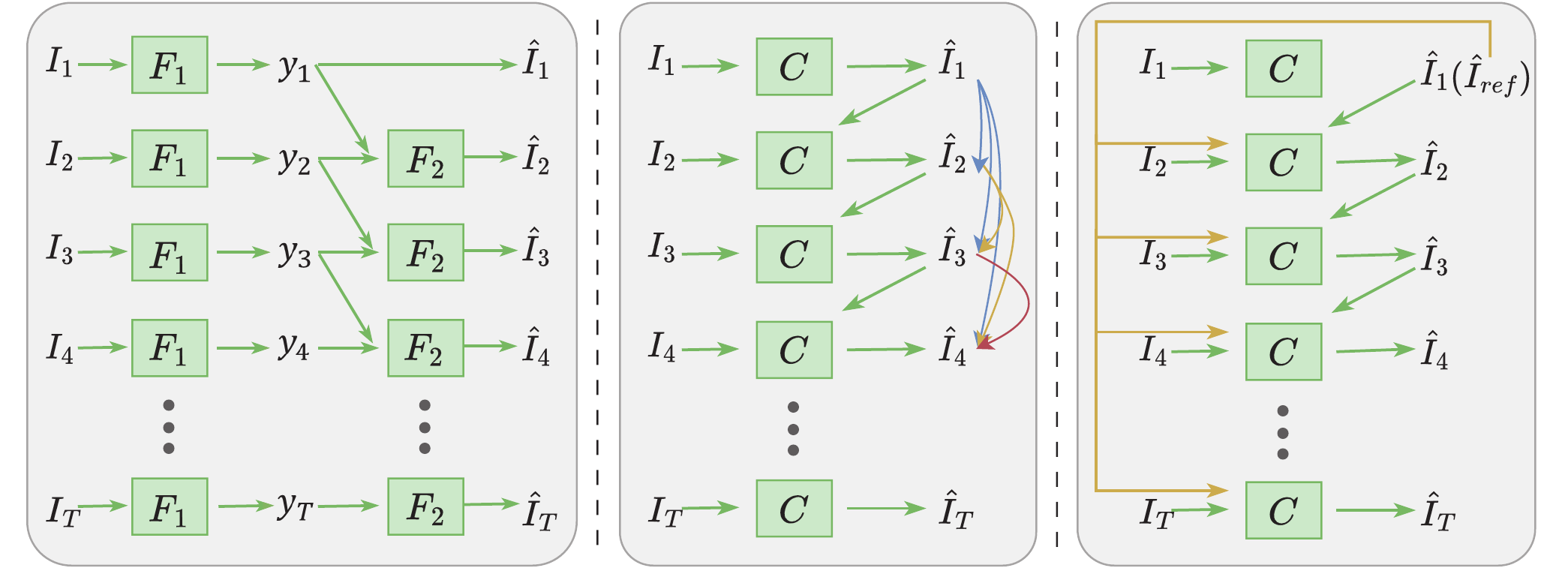}
   \caption{Comparison of different frameworks in video colorization: (a) Colorization with post-processing~\cite{Blind,LearningBlind,TemporallyC,DeepPrior}, (b) Colorization with dense long-term loss~\cite{VCGAN}, (c) Colorization with semantic correspondence (Ours), where $I$ is the input, $r$ is the reference image, $\hat I$ is the video colorization result and $y$ is the image colorization result. $F_1,F_2,C$ represent CNNs, and the color lines in (b) indicate dense long-term loss.}
   \label{fig:compare}
\end{figure*}
\label{sec:intro}
As a well-known ill-posed problem, video colorization task owns serious ambiguity that a grayscale object could be plausible in various colors. This characteristic usually results in temporal inconsistency that the colors of an object may change in different frames. In order to resolve such inconsistency, mainly three kinds of methods are proposed: task-independent, fully-automatic and exemplar-based methods.

The task-independent~\cite{Blind,LearningBlind,TemporallyC,DeepPrior} methods aim to enhance temporal consistency between image colorization results via post-processing. They formulate a temporal filter and punish the warping errors computed by optical flow between adjacent frames. However, their results are still not consistent enough when the generated colors are extremely different in adjacent frames, and they have to process each frame twice. Based on the conception of task-independent methods, automatic colorization methods \cite{FullyAuto,automatic3D,AutomaticTC,VCGAN} are proposed. They directly map the feature embedding of grayscale images to their color representations by learning from large datasets. For instance, Lei et al.~\cite{FullyAuto} divide the video colorization into a single frame colorization subnet and a smoothing subnet. However, it is difficult to generate colorful results. To integrate both image and video colorization, Zhao et al.~\cite{VCGAN} propose an end-to-end network using two step training, and introduce a dense long-term loss to minimize flickers of generated frames. However, the long-term loss only covers few frames and is dependent on the quality of optical flow. For long videos, it still suffers temporal inconsistency in wide frame interval. \cref{fig:compare} illustrates the different colorization strategy.

The exemplar-based methods utilize a colorized reference image to supervise the colorization process for all frames~\cite{VPN,trackingbycolor,Switchable,Deepremaster,DeepExamplar,referenceM}. A semantic correspondence network is usually adopted to find the pixel-wise correspondence between reference and grayscale images. For example, Zhang et al.~\cite{DeepExamplar} propose a recurrent network where the non-local operation~\cite{non-local} is responsible to find semantic correspondence between reference and grayscale images, and the previous colored frame is also leveraged to increase temporal consistency. To further enhance spatiotemporal long-term dependency in videos, Chen et al.~\cite{exemplarvcld} propose a  double-head non-local operation and an attention~\cite{attention} based linkage subnet to improve the representation ability. However, the behavior of exemplar-based method is highly dependent on the selection of reference image, and the manual selection of reference is usually experience required and time-consuming.

Under this circumstance, this paper proposes Temporal Consistent Automatic Video Colorization with Semantic Correspondence, which combines semantic correspondence network into automatic video colorization. As difficult for automatic methods to keep long-range consistency, a reference image together with semantic correspondence network is leveraged to supervise the whole colorization process; and as complicated to manually select the reference image, a prior reference colorization network is leveraged to generate the reference image by automatically coloring the first frame in video. Such direct colorization of reference can not only avoid manual selection, but also increase the similarity of reference and grayscale images (since they belongs to the same video), which is beneficial for semantic correspondence.
Our contributions can be summarized as:
\begin{itemize}
  \item A novel framework combines automatic video colorization with semantic correspondence is proposed to keep long-range consistency.
  \item We leverage an automatically generated reference image to supervise the colorization of remaining frames. Each frame is supervised by both the reference image and the immediately colorized preceding frame.
  \item Experiments demonstrate that our method can better maintain temporal consistency, and outperforms recent state-of-the-arts both qualitatively and quantitatively. 
\end{itemize}

\section{Related Works}
In this section, we will introduce the two main methods
in video colorization: automatic and
exemplar-based.

\subsection{Automatic Colorization}
Automatic methods \cite{FullyAuto,automatic3D,AutomaticTC,VCGAN} are proposed to further optimise temporal coherence. They map grayscale images directly to color embedding using deep neural networks, while maintaining frame continuity. 
Lei et al.~\cite{FullyAuto} propose a multi-modal automatic framework that can generate four diverse colorization results simultaneously. To maintain spatio-temporal consistency, they impose similarity between pixel pairs by K Nearest Neighbor (KNN) search in feature space or by optical flows. Zhao et al.~\cite{VCGAN} propose a hybrid recurrent network that integrates both image and video colorization and meanwhile leverage a dense long-term loss which considers not only adjacent but long-term continuity to optimize it. Nevertheless, it is as yet difficult to generate a colorful result with the help of these methods. Especially in practical applications like old movie restoration, there are certain colors in specific scenarios for objects such as clothes, skin, house, which have historical basis and are difficult to generate by fully-automatic approaches.
\begin{figure*}[!t]
   \centering
   \includegraphics[width=6in]{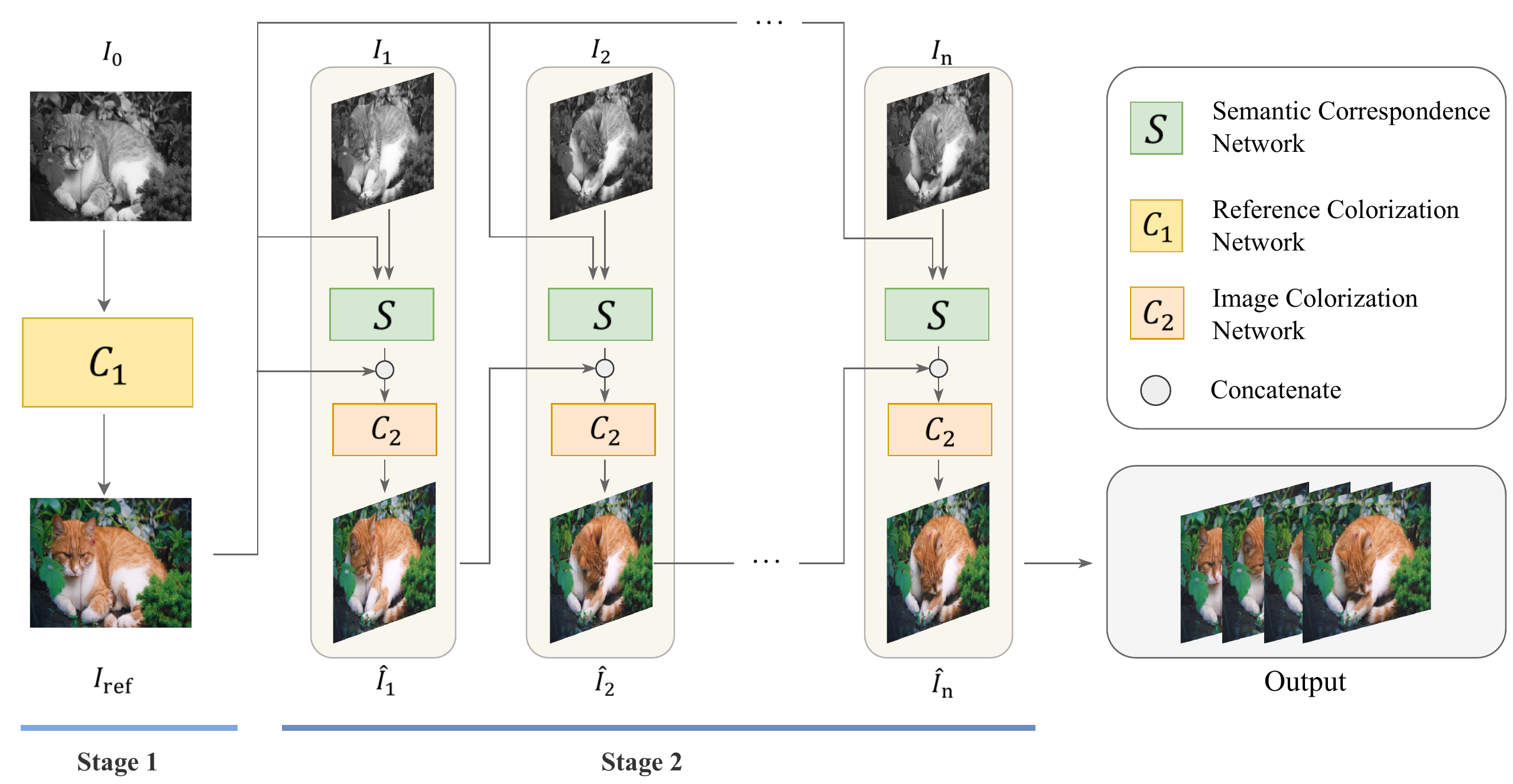}
   \caption{The overall framework of our method. There are mainly three components: a reference colorization network, an image colorization network and a semantic correspondence network. The reference colorization network generate a colorized reference image using the first grayscale frame of the video. The semantic correspondence network and the image colorization network then leverage the reference to supervise the whole colorization process.}
   \label{fig:vectoria_overall}
\end{figure*}
\subsection{Exemplar-based Colorization}
Exemplar-based methods generally utilize one or more colored frames in a video as reference images to guide the colorization process. These method \cite{jacob2009colorization,ben2015approximate,xia2016robust} leverage handcrafted low-level features to find temporal correspondence between frames and colorize following frames in sequence. More recent methods tend to use deep neural networks to achieve temporal propagation \cite{VPN,trackingbycolor,Switchable}. While these approaches produce much more colorful results, their coloration depends only on the previous frame, which makes it easy to accumulate color errors as they propagate. Another type of method involves the reference images throughout the process \cite{Deepremaster,DeepExamplar,referenceM}, providing more stable results. For instance, Zhang et al.~\cite{DeepExamplar} propose a recurrent framework with novel loss functions where colorization depends on both the reference and the previous frames. Iizuka et al.~\cite{Deepremaster} first propose a single framework for remastering vintage films. They adopt a source-reference attention that can handle multiple references, and utilize 3D-CNN for modeling temporal correspondence. Although favorable results are obtained, these approaches nonetheless lack long-term spatio temporal dependencies, likely to wash out color in motion areas. Different from previous methods, our method has strong ability in modeling long-term dependency both spatially and temporally.

\section{Method}
\subsection{Problem formulation}
Given consistent grayscale video frames $\{I^l_1,I^l_2,...,I^l_n\}$, the colorization task aims to generate corresponding colorized frames  $\{\hat I^{lab}_1,\hat I^{lab}_2,...,\hat I^{lab}_n\}$, where $l$ and $ab$ denote the luminance and chrominance in CIELAB color space respectively. On the one hand, the generated result $\hat I^{lab}_n$ should be perceptually similar to the ground truth image $I^{lab}_n$. On the other hand, the current frame $\hat I^{lab}_n$ should not only be temporal consistent to its adjacent frames $\hat I^{lab}_{n-1},\hat I^{lab}_{n+1}$, but also be similar to the frames with wide temporal interval (e.g. $\hat I^{lab}_1$). For recent automatic colorization methods~\cite{FullyAuto,VCGAN,AutomaticTC}, the colorization of $I^{l}_n$ usually based on the previously colorized frame:
\begin{equation}
  \label{vectoria:auto}
   \hat I_n^{lab} = \mathcal{F}_{auto}(I_{n}^l,\hat I_{n-1}^{lab}),
\end{equation}
where $\mathcal{F}_{auto}$ denotes the automatic colorization network. Such methods colorize the video in manner of a Markov Chain, while the consistency is established only for adjacent frame, and the frames in wide interval may be inconsistent. Meanwhile, exemplar-based methods~\cite{DeepExamplar,exemplarvcld} usually colorize a frame depending on an additional reference image $I_{ref}$:
\begin{equation}
  \label{vectoria:auto}
   \hat I_n^{lab} = \mathcal{F}_{exemp}(I_{n}^l,\hat I_{n-1}^{lab},I_{ref}),
\end{equation}
where $\mathcal{F}_{exemp}$ represents the exemplar-based video colorization network. The reference image is responsible to supervise the colorization process. It determines the color style of images, thus reduce color ambiguity and enhance temporal consistency. However, the reference image usually needs manual selection which is experience required and time-consuming. Therefore, this paper propose a two-stage colorization framework where  the reference image is automatically generated and supervise the colorization.

\subsection{Two-stages Colorization}
  Our overall framework is illustrated in  \cref{fig:vectoria_overall}. The framework is divided into two stages. The first stage involves an automatic reference colorization network, and the second stage includes a semantic correspondence network and an image colorization network. In the first stage, the first frame of each videos is selected to be automatically colorized. And the resulting image is then regarded as the reference image in the second stage. 
 \begin{equation}
  \label{vectoria:stage_1}
   \hat I_{ref}^{lab} = \mathcal{C}_1(I_0^l),
 \end{equation}
 where $\mathcal{C}_1$ represents the reference colorization network. $I_i$, $\hat I_{ref}$ denote the $i-th$ frame and the reference image respectively. For maintaining temporal consistency, rather than only correlated to the previous few frames, the colorization of the remaining grayscale frames also depends on their semantic correspondence with the reference image, which can be formulated as:
 \begin{equation}
  \label{vectoria:stage_2}
   \hat I_n^{lab} = \mathcal{C}_2(\mathcal{S}(I_{n}^l,\hat I_{ref}^{lab}),\hat I_{n-1}^{lab}),
 \end{equation}
 where $\mathcal{S}$ represents the semantic correspondence network, and $\mathcal{C}_2$ the image colorization network in the second stage. Thus, our approach is capable of better maintaining temporal consistency along time series. 

\subsection{Loss function} 
As an inherent ambiguous problem, it is improper to directly compare the color difference between the ground truth and generated image. Recently, the perceptual difference has been proved to be robust to appearance differences caused by two plausible colors~\cite{DeepExamplarimage}. It compares the difference between features $reluL\_2$  extracted by pretrained VGG-19 network~\cite{vgg}. In this paper, the coarse-to-fine perceptual loss is leveraged:
\begin{equation}
  \label{perc}
  \mathcal L_{perc} = \sum_L \alpha_L \| \Phi_L(\hat T) - \Phi_L(T) \|_2^2 ,
\end{equation}
here $L\in\{ 3,4,5 \}$, and $\alpha_L\in\{0.02,0.003,0.5\}$ denotes corresponding weight coefficient. The coarse-to-fine strategy involves the comparison of both high-level and low-level feature representations.

Besides, we empirically find that the $L_1$ loss helps the convergence of network, and the smooth loss~\cite{DeepExamplar} helps to reduce color bleeding. Moreover, the PatchGAN~\cite{pix2pix} is also adopted to increase high-frequency color fidelity. It classifies each patch as real or fake rather than the whole image. For networks in the first stage, the overall objective loss can be written as:
\begin{equation}
  \label{loss_all}
  \begin{aligned}
  \mathcal L_1 = & \lambda_{perc}L_{perc}+\lambda_{L_1}L_{L_1}+\lambda_{smooth}L_{smooth} \\ & +\lambda_{patch}L_{patch}
  \end{aligned}
\end{equation}
For networks in the second stage, the temporal warping loss~\cite{TemporallyC} is further adopted to constraint temporal consistency. The corresponding objective loss is:
\begin{equation}
  \label{loss_all}
  \begin{aligned}
  \mathcal L_2 = & \lambda_{perc}L_{perc}+\lambda_{L_1}L_{L_1}+\lambda_{smooth}L_{smooth} \\ & +\lambda_{patch}L_{patch} +\lambda_{temp}L_{temp}
  \end{aligned}
\end{equation}

 \begin{figure}[!t]
   \centering
   \includegraphics[width=\linewidth]{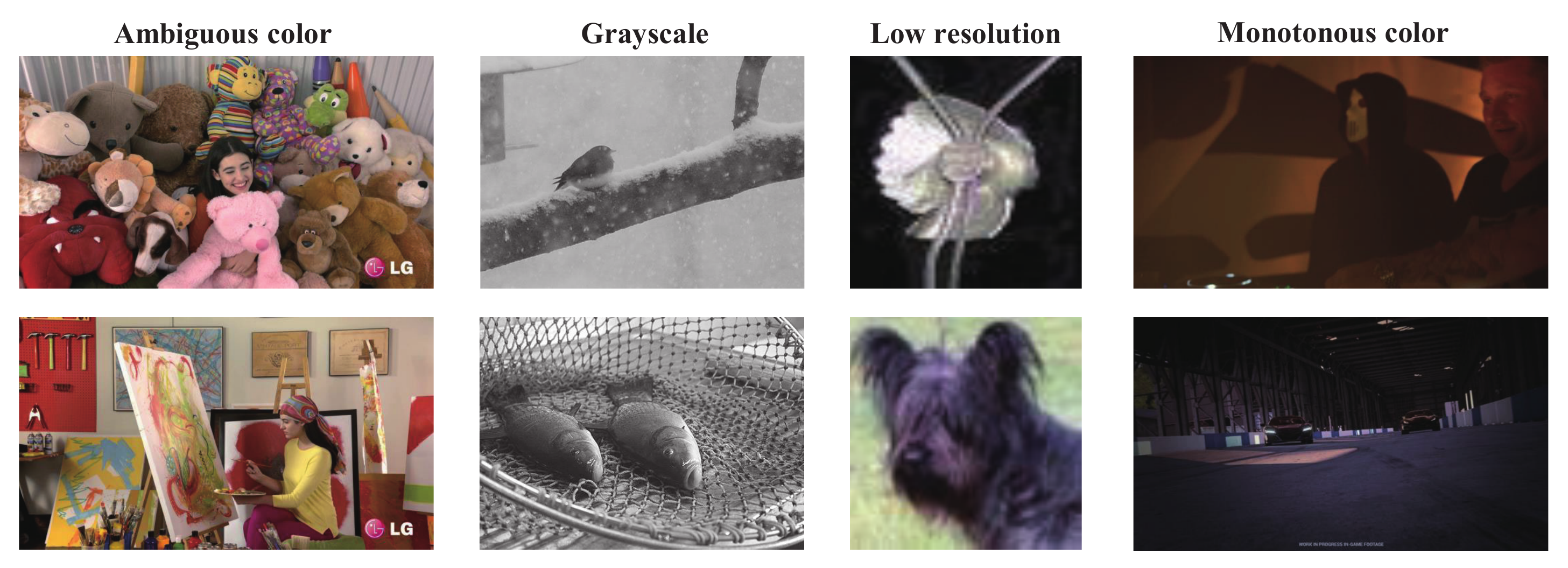}
   \caption{Examples of bad images removed from our training set. Including images with ambiguous colors, monotonous colors, low resolution or grayscale.}
   \label{fig:badimgs}
\end{figure}

\section{Implementation}
{\bf Network Structure.} The reference colorization network is an encoder-decoder structure with skip connections, group convolutions and dilated convolutions~\cite{dilatedconv}. The semantic correspondence network is a CNN-Transformer structure~\cite{conformer} with non-local operation~\cite{non-local}. And the image colorization network combines the encoder-decoder structure in the first stage with a Transformer branch. The network structure in the second stage is basically the same as in~\cite{exemplarvcld}, and we recommend to check out more details from the original paper.

 {\bf Training. }The training process of the networks in two stages is independent. For network in the first stage, the reference colorization network is trained on images from ImageNet~\cite{ImageNet}, REDS~\cite{reds}, DAVIS~\cite{davis}, SportMOT~\cite{sportsmot} and the official training set in NTIRE 2023 Video Colorization Challenge~\cite{ntire23videocolorization}. We remove the images with ambiguous colors, monotonous colors, low resolution or grayscale (\cref{fig:badimgs}). About 1.1 million of images are involved in training. For networks in the second stage, the training set includes DAVIS~\cite{davis}, Videvo~\cite{videvo} and FVI~\cite{fvi} dataset. 2090 videos in total are collected. And we train the networks in manner of frame propagation (i.e. the first frame in each video is regarded as the reference image). Moreover, The pretrained models in \cite{DeepExamplar,Vistr} are used to initialize the parameters.  One can refer to our published code for more implementation details.

\begin{figure*}[!t]
   \centering
   \includegraphics[width=\linewidth]{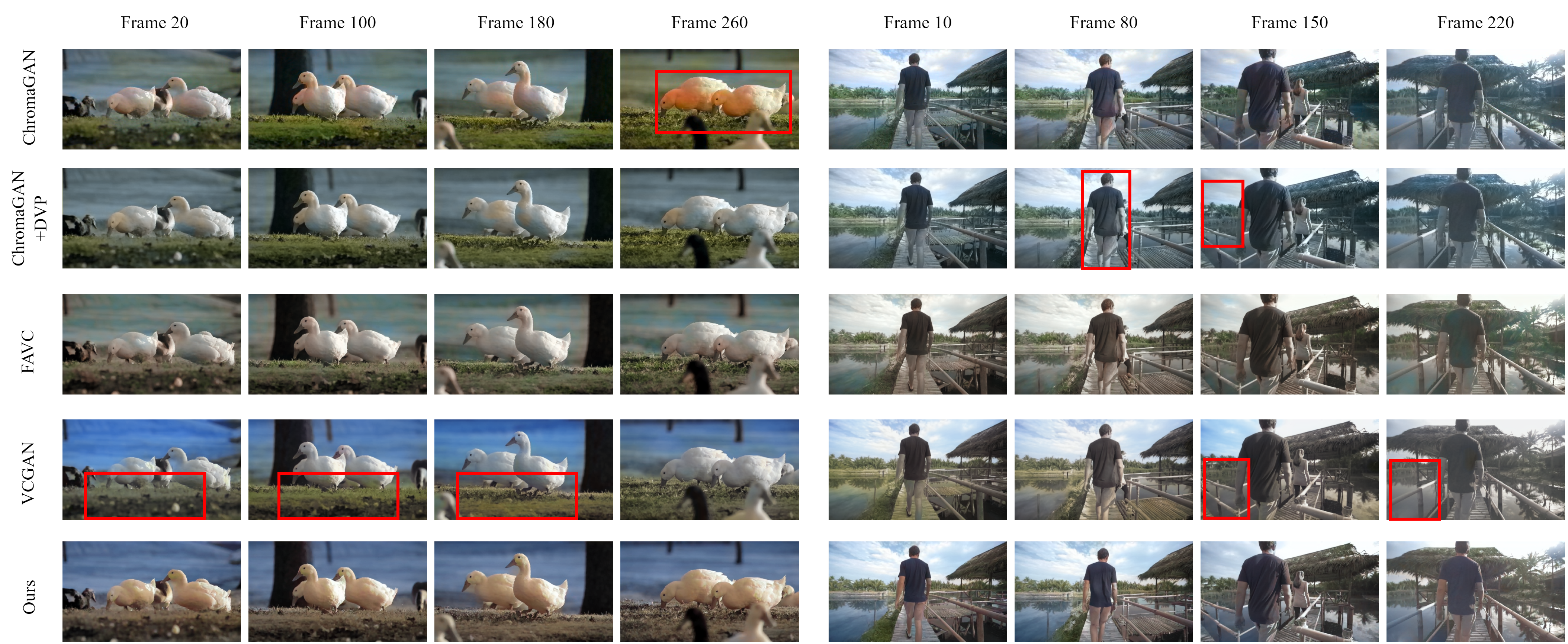}
   \caption{Visual comparison with the state-of-the-art methods on Videvo test set. From top to bottom are the methods of ChromaGAN~\cite{Chromagan}, ChromaGAN+DVP~\cite{DeepPrior}, FAVC~\cite{FullyAuto} and VCGAN~\cite{VCGAN} respectively. Our method achieves the most colorful while consistent result.}
   \label{fig:compare}
\end{figure*}


\begin{table*}[!t]
\centering
\caption{Quantitative comparison with state-of-the-art methods on DAVIS and Videvo dataset. Our method gets the best FID, while maintains comparable CDC.}
\label{tab:compare_sota}
\begin{tabular}{lccccc}
\toprule
              & \multicolumn{2}{c}{DAVIS}          & \multicolumn{2}{c}{Videvo}         &                         \\
Method        & FID$\downarrow$     & CDC$\downarrow$    & FID$\downarrow$     & CDC$\downarrow$      & Model type              \\ \midrule
ChromaGAN~\cite{Chromagan}     & 52.97          & 0.008771          & 50.57          & 0.004565          & Fully-automatic (image) \\
ChromaGAN+DVP~\cite{DeepPrior} & 58.94          & \textbf{0.003672} & 58.85          & 0.001967          & Task-independent        \\
FAVC~\cite{FullyAuto}          & 58.33          & 0.003682          & 57.08          & \textbf{0.001575} & Fully-automatic         \\
VCGAN~\cite{VCGAN}         & 59.58          & 0.008951          & 67.48          & 0.003208          & Fully-automatic         \\
Ours          & \textbf{46.28} & 0.003836          & \textbf{49.02} & 0.001681          & Fully-automatic         \\ \bottomrule
\end{tabular}
\end{table*}

\section{Experiment}
\subsection{Comparisons with state-of-the-arts}
In this section, state-of-the-art methods including ChromaGAN~\cite{Chromagan}, DVP~\cite{DeepPrior}, FAVC~\cite{FullyAuto} and VCGAN~\cite{VCGAN} are compared with our method both quantitatively and qualitatively. The official published code of the methods are used for comparison.

\begin{figure*}[!t]
   \centering
   \includegraphics[width=\linewidth]{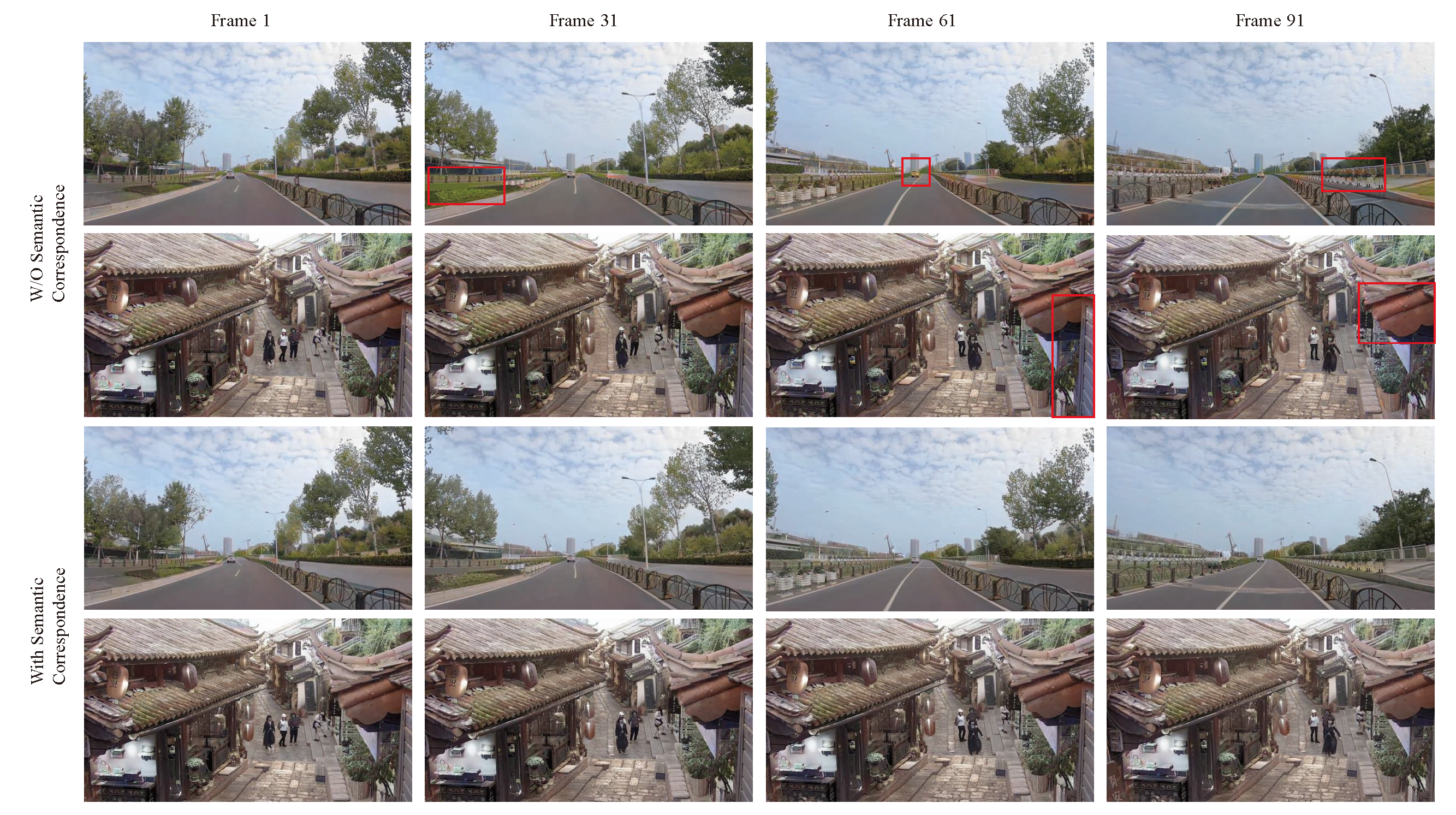}
   \caption{Visual comparison of colorization results for networks with or without semantic correspondence on the test set of NTIRE 2023 Video Colorization Challenge~\cite{ntire23videocolorization}. Each interval of the adjacent frames is 30.}
   \label{fig:vectoria_example}
\end{figure*}

\begin{table*}[!t]
\centering
\caption{Comparison of our method with task-independent method DVP~\cite{DeepPrior} on DAVIS test set. Our method maintains better color perceptual quality while obtaining the better temporal consistency compared to DVP. Moreover, we try to combine DVP with our method (the last row), and the CDC further improved, but with apparent performance drop in FID.}
\label{tab:ablation_corr}
\begin{tabular}{ccccccc}
\toprule
\begin{tabular}[c]{@{}c@{}}Image\\ colorization\end{tabular} &
  \begin{tabular}[c]{@{}c@{}}Semantic\\ correspondence\end{tabular} & 
  DVP &
  FID$\downarrow$ &
  CDC$\downarrow$ &
  Time (s) &
  Parameters (M) \\ \midrule
\checkmark &           &           & 41.12 & 0.005045 & 0.4648        & 32.80        \\
\checkmark &           & \checkmark & 58.85 & 0.004189 & 0.4648+0.9884 & 32.80+23.93 \\
\checkmark & \checkmark &           & 46.28 & 0.003836 & 0.4718        & 32.82+148.21  \\  
\checkmark & \checkmark & \checkmark & 59.92 & 0.003708 & 0.4718+0.9884  & 32.82+23.93+148.21  \\ \bottomrule
\end{tabular}
\end{table*}
{\bf Quantitative comparison.} For quantitative comparison, the image quality metric FID (Fr´echet Inception Distance)~\cite{fid} and temporal metric CDC (Color Distribution Consistency index)~\cite{TemporallyC}  are adopted, as which are widely used in previous works~\cite{DeepExamplarimage,DeepExamplar,exemplarvcld,ntire23videocolorization,TemporallyC}. The FID measures the semantic distance between generated and ground truth images. The lower the FID, the more natural the image result. And the CDC computes the Jensen-Shannon (JS) divergence of the color distribution between consecutive frames. More consistent video will get lower CDC. We experiment on the test set of DAVIS~\cite{davis} and Videvo~\cite{videvo} dataset, the quantitative result is illustrated in \cref{tab:compare_sota}. The ChromaGAN gets excellent FID, but with bad CDC since it is an image colorization method without temporal modeling. With DVP, the CDC of ChromaGAN obviously declines, but its FID gets worse at the same time. FAVC achieves the second best and the best CDC in the two datasets respectively. However, it gets high FID. In both of the two datasets, our method achieves the best FID, while maintains comparable CDC with the best results.

{\bf Qualitative comparison.}  The visual comparison of the methods on Videvo test set is illustrated in \cref{fig:compare}. ChromaGAN generates colorful result, but the object color can be very different from frame to frame, as in the case of the ducks in the left video. The DVP distinctly removes the temporal inconsistency of ChromaGAN. However, it also washes out the colors in images (like in the right video), as it is tend to remove the bright but inconsistent colors rather than propagate the bright colors to other frames. The result of FAVC is quite consistent, but it is not colorful enough compared with other methods. VCGAN models dense temporal consistency in small frame interval. But with large frame interval, distinct inconsistency can be observed. Such as the grass in the left video and the water in the right video. Though the dense consistency in VCGAN with large interval is feasible, it requires huge computational consumption that dozens or hundreds of optical flows are required for a single image's training. With semantic correspondence, our method achieves the most colorful while consistent result.

\begin{figure*}[!t]
   \centering
   \includegraphics[width=\linewidth]{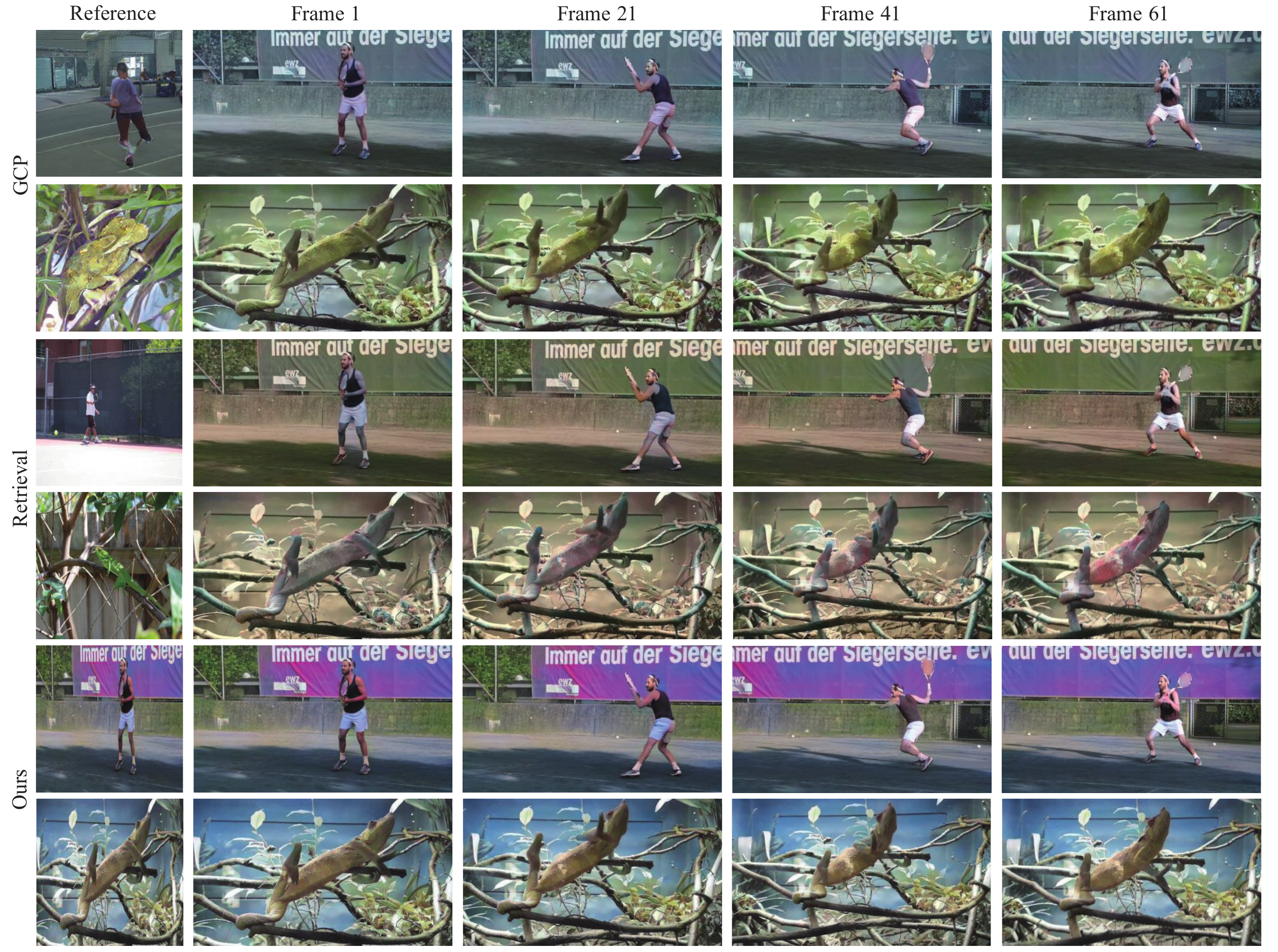}
   \caption{Comparison of the colorization via reference images obtained by different methods on DAVIS test set. The methods including GCP (Generative Color Prior)~\cite{GCP}, a retrieval method~\cite{DeepExamplarimage} and our reference colorization network. Each interval of the adjacent frames is 20. Our method obtains more colorful and realistic results.}
   \label{fig:diff_ref}
\end{figure*}

\begin{table}[!t]
\centering
\caption{Comparison of different methods to automatically obtain reference image on DAVIS test set. Note that we have not specially trained the stage 2 network for reference images in stage 1.}
\label{tab:ablation_ref}
\begin{tabular}{lcccc}
\toprule
\multicolumn{1}{l}{} & \multicolumn{2}{c}{DAVIS} & \multicolumn{2}{c}{Videvo} \\
Reference            & FID$\downarrow$        & CDC$\downarrow$          & FID$\downarrow$        & CDC$\downarrow$           \\ \midrule
Retrieval~\cite{DeepExamplarimage}            & 53.38      & 0.003939     & 48.24      & 0.001793      \\
GCP~\cite{GCP}                  & 50.19      & 0.003843     & 48.51      & 0.001710      \\
Ours-r                 & 46.17      & 0.004179     & 45.35      & 0.001684      \\ \bottomrule
\end{tabular}
\end{table}

\subsection{Ablation study}
{\bf Effect of semantic correspondence.} We train another model without the semantic correspondence network in order to represent its effectiveness, and the task-independent method DVP~\cite{DeepPrior} is also adopted for comparison. The quantitative result is reported in \cref{tab:ablation_corr}. Combining DVP with image colorization, the CDC gets improved, but with huge decline of FID (from 41.12 to 58.85). While with semantic correspondence, our method achieves better CDC and less drop in FID (from 41.12 to 46.28), which represents that our method maintains better color perceptual quality while obtaining better temporal consistency.  Moreover, we try to combine DVP with our method, and the CDC further improved, but with apparent performance drop in FID (from 46.28 to 59.92).

Beside, the processing time per image and number of network parameters in \cref{tab:ablation_corr} represents that our method consumes less time than task-independent method DVP (0.4718 sec. compared to 0.4648+0.9884 sec. per image) though with more parameters (32.82+148.21 M compared to 32.80+23+93 M). Moreover, the visual comparison for networks with or without semantic correspondence is illustrated in  \cref{fig:vectoria_example}. Without semantic correspondence network, the object can have diverse colors in different frames (e.g. the color of the car could be white in frame 1  and yellow in frame 61). While with the semantic correspondence network, the frames with large interval still maintain pleasant temporal consistency.

{\bf Different reference selection strategy.} We further introduce two methods to automatically obtain the reference image for video colorization, which are: (1) GCP (Generative Color Prior)~\cite{GCP}, a generation method based on BigGAN~\cite{biggan}. The GCP learns a mapping from a grayscale image to a embedding which acts as the condition of BigGAN to generate a colorized image similar with the grayscale image. (2) a retrieval method~\cite{DeepExamplarimage} using PCA-based compression~\cite{pca-retrieval}. This method compares the PCA embedding of the grayscale image with embeddings of images from large dataset (e.g. ImageNet). And the image with largest correlation will be selected. The two methods are compared with our reference colorization network on DAVIS and Videvo datasets. The reference image obtained by three methods are used in our stage 2 network respectively. As the stage 2 networks are trained by frame propogation, we train another network mostly leveraging the retrieved images as references like in~\cite{DeepExamplar}. As shown in \cref{tab:ablation_ref}, in DAVIS dataset, our method gets the best FID. And in Videvo dataset, our method obtains both the best FID and CDC. Besides, the visual comparison is shown in \cref{fig:diff_ref}. It can be noticed that our results are more colorful and realistic. We believe this is because our method obtains a reference image that is more similar to the grayscale frames, so we are able to transfer colors more precisely and obtain more vivid results.

\begin{table}[!t]
\centering
\caption{Test results on NTIRE 2023 Video Colorization Challenge~\cite{ntire23videocolorization}. Our method obtains the 3rd place in CDC track with 52.68\% improvement over the baseline.}
\label{tab:ntire2023}
\begin{tabular}{lcc}
\toprule
Team         & FID$\downarrow$   & CDC$\downarrow$    \\ \midrule
MiAlgo       & 54.72 & 0.000819 \\
CUCPLUS      & 26.79 & 0.000962 \\
\textbf{Ours} & \textbf{63.76} & \textbf{0.001017} \\
NJUSTer      & 62.45                     & 0.001066                     \\
ppzz         & 56.81                     & 0.001122                     \\
LVGroup HFUT & 63.71                     & 0.001525                     \\
baseline     & 61.30                     & 0.002149                     \\ \bottomrule
\end{tabular}
\end{table}

\subsection{NTIRE 2023 Challenge}
We have proposed our method for NTIRE 2023 Video Colorization Challenge~\cite{ntire23videocolorization}. Our entry obtains the 3rd place in Track 2: Color Distribution Consistency (CDC) Optimization and the CDC score is very close to the second method. The goal of this track is to obtain the best CDC result while being constrained to maintain FID.
The benchmark results of our model and the other teams in NTIRE 2023 are shown in \cref{tab:ntire2023}.

\subsection{Limitations}
Despite the promising progress of our method for maintaining video temporal consistency, there are still some limitations.

Since we include the semantic correspondence network, our method is not robust enough when the scene changes in videos, which is an inherent weakness of the exemplar-based video colorization~\cite{exemplarvcld}. Meanwhile, as the reference image is automatically colorized, it is not likely to generate diverse image results. 

\begin{figure}[!t]
   \centering
   \includegraphics[width=\linewidth]{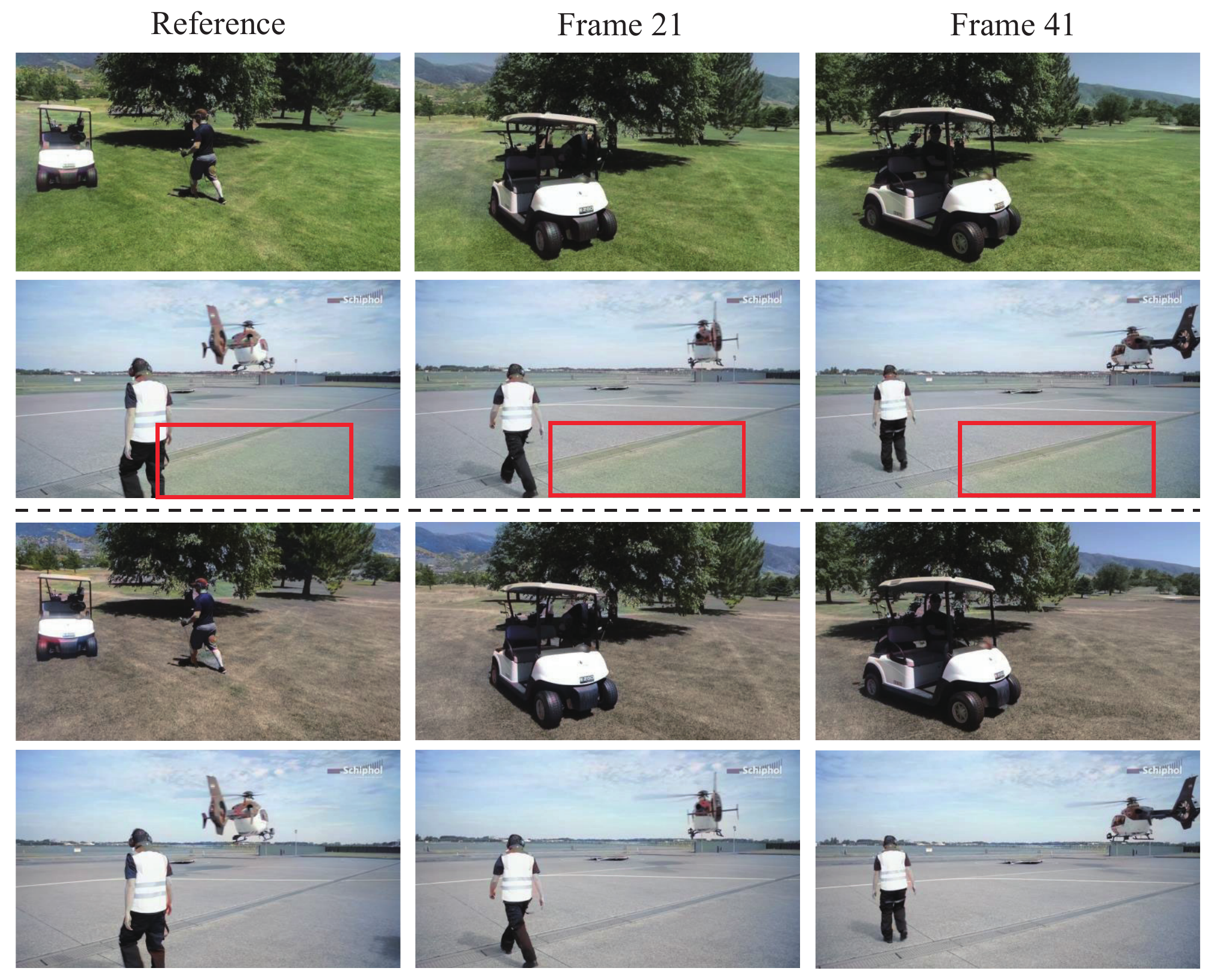}
   \caption{Colorization with different reference images obtained from our reference colorization network. The first and the second row show the results of the original model. The third and the fourth row show the results of another model trained without ImageNet~\cite{ImageNet} dataset. }
   \label{fig:first_quality}
\end{figure}

Moreover, the performance of the final colorization result is highly dependent on the quality of the reference image. As shown in \cref{fig:first_quality}, we can apparently observe that the reference image has a high similarity to the subsequent colored images in terms of the color style. The reason is that the most of the colors from the reference image are considered plausible and will be transferred directly to the grayscale images. Even the incorrectly colored regions may still be considered as supervision information to guide the colorization of subsequent grayscale images. That is, the colorization of the reference image will affect the style of the videos (e.g., the color styles of the first and third rows are quite different due to the different reference images.), or even lead to unexpected colors (e.g., the ground in the second row is colorized to unpleasant green).

\section{Conclusion}
In this paper, we propose a novel automatic video colorization method via semantic correspondence, which utilize an automatically generated reference image to supervise the colorization process and preserve temporal consistency. Our intuition is to fully exploit the semantic correspondence between video frames to improve the colorization consistency of the network. Experiment also demonstrated that our method is capable of better maintaining color consistency in large frame interval than recent methods. Finally, ablation studies show the effectiveness of the network components.

\subsection*{Acknowledgements} This work was supported by the National Key R\&D Program of China (2021ZD0109802) and the National Natural Science Foundation of China (81972248).

{\small
\bibliographystyle{ieee_fullname}
\bibliography{egbib}
}

\end{document}